\documentclass{article} 
\usepackage{iclr2021_conference,times}

\usepackage{amsmath,amsfonts,bm}

\def\eqref#1{equation~\ref{#1}}

\def\1{\bm{1}}

\DeclareMathAlphabet{\mathsfit}{\encodingdefault}{\sfdefault}{m}{sl}
\SetMathAlphabet{\mathsfit}{bold}{\encodingdefault}{\sfdefault}{bx}{n}

\DeclareMathOperator*{\argmax}{arg\,max}

\usepackage{hyperref}
\usepackage{url}
\usepackage{times}
\usepackage{latexsym}
\usepackage{appendix}

\usepackage{graphicx}
\usepackage{todonotes}
\usepackage{subcaption}
\usepackage{makecell}
\usepackage{multirow}
\usepackage{todonotes}
\usepackage{amsmath}
\usepackage{enumitem}
\usepackage{hyperref}
\usepackage{pgfplots}
\usepackage{caption}
\usepackage{algorithm}
\usepackage{algpseudocode}
\usepackage{tabularx,ragged2e}

\usepackage{microtype}

\algnewcommand{\LineComment}[1]{\State \(\triangleright\) #1}
\newcommand\bh{\mathbf{h}}
\newcommand\ba{\mathbf{a}}
\newcommand\bH{\mathbf{H}}

\title{Improving Longer-Range Dialogue State Tracking}

\author{Ye Zhang, Yuan Cao, Mahdis Mahdieh, Jeffrey Zhao, Yonghui Wu 
\\
Google Research\\
\texttt{\{yezhan,yuancao,mahdis,jeffreyzhao,yonghui\}@google.com} \\
}

\iclrfinalcopy 
\begin{document}

\maketitle

\begin{abstract}
Dialogue state tracking (DST) is a pivotal component in task-oriented dialogue systems. While it is relatively easy for a DST model to capture belief states in short conversations, the task of DST becomes more challenging as the length of a dialogue increases due to the injection of more distracting contexts. In this paper, we aim to improve the overall performance of DST with a special focus on handling longer dialogues. We tackle this problem from three perspectives: 1) A model designed to enable hierarchical slot status prediction; 2) Balanced training procedure for generic and task-specific language understanding; 3) Data perturbation which enhances the model's ability in handling longer conversations. We conduct experiments on the MultiWOZ benchmark, and demonstrate the effectiveness of each component via a set of ablation tests, especially on longer conversations.
\end{abstract}

\section{Introduction}
Dialog state tracking (DST) is a key component in modern task-oriented dialogue (ToD) systems. The goal of DST is to keep track of a user's requirements from the conversation between the user and agent, it is therefore critical for a DST model to be accurate in reasoning about the belief states from dialogue history, especially when the dialogue length increases as the conversation further unfolds.

While significant progress has been made in recent years on DST~\citep{chao2019bert,wu2019transferable,zhang2019find,mehri2020dialoglue,hosseini2020simple,li2020coco}, mostly driven by the advancements in NLP techniques including pre-training followed by task-specific fine-tuning ~\citep{devlin2018bert,liu2019roberta} and large-scale modeling ~\citep{radford2019language,brown2020language,raffel2020exploring}, it remains a challenging problem how to properly handle multi-turn, multi-domain dialogues with long contexts. Although a BERT-style pre-trained encoder is a standard component in many of the state-of-the-art DST models and is capable of making precise belief state predictions when the context is short and clean, it can still be easily misguided by distracting information from remote turns hence fails to reason about the current states.

To further improve the DST model quality, especially for dialogues with many turns and covering multiple domains, we believe that one should not only improve the model's generic and task-specific language understanding ability, but also properly design the model such that it becomes more selective about the information it needs to predict and guards itself against distractions from prolonged dialogue contexts. Following this line of thought, in this paper we investigate the following ways of improvement: 1) Hierarchical prediction of slot status; 2) Training procedure that balances generic and DST-specific language understanding; 3) Data perturbation which enhances a model's ability of handling longer conversations. Our experimental results on the MultiWOZ benchmark~\citep{budzianowski2018large,eric2019multiwoz} show that not only do these techniques lift the overall accuracy of state tracking, but they are mostly effective in improving longer conversations while the progress made for short ones slow down. Nevertheless, there is still much room for improvement for long-range conversations, which deserves more focus for future research work.
 
\section{Model}
\subsection{Base model} \label{sec:base_model}
Our base model is built upon a pre-trained BERT encoder as backbone, on top of which multiple prediction heads are mounted to predict state values and their statuses. Inputs to the model are the conversation histories concatenated with the most recent utterance. Following a similar strategy as ~\cite{zhang2019find}, we divide states into two types, namely categorical and non-categorical slots. Categorical slots are those having a pre-defined, closed-set of candidate values, such as ``cheap'' and ``expensive'' for restaurant price range. Non-categorical slots are those whose values cannot be practically pre-defined and better be extracted directly from utterances, for example the departure time of a train.

Upon receiving each new utterance, the model first encodes the history together with the utterance. The same encoder is also used to encode the names of each candidate categorical and non-categorical slots. The combination of the representations of candidate slots and utterances are then used to make predictions for the slot values: for categorical slots their categories are predicted by a softmax, and for non-categorical slots sigmoid functions are used to indicate the start and end positions of value spans in the utterances. For each candidate slot, a softmax is also used to predict the state status, which is one of \{\verb|active, don't care, inactive|\}. The predicted values are only considered for those slots whose status are inferred as \verb|active|.

\subsection{Hierarchical Slot Status prediction} \label{sec:hier_gate}
One problem with the DST model is that the state prediction accuracy heavily hinges on the correctness of the slot status. Rightly predicted slot values are immediately ignored as long as the slots are assigned wrong statuses. In fact, even with the simplest setup of our baseline models, which reaches 48.1\% joint-goal-accuracy (JGA) on the MultiWOZ 2.1~\citep{eric2019multiwoz} benchmark, if we do a cheating experiment and use oracle slot statuses instead of their predictions during inference, the JGA immediately rises up to 85.75\%! This reveals the criticality of making correct predictions for slot status.

Status prediction is difficult as a majority of slots are inactive at any turn in a conversation, which creates a severe class imbalance problem for both training and inference procedures. What is more, it becomes even less accurate as the length of a conversation increases, due to the introduction of more distracting information. To mitigate these issues, we propose a hierarchical slot status prediction procedure. We first predict the domain of the current turn, then only predicts status of slots relevant to the active domain. This enables us to narrow down the range of slots to a small subset most likely to be relevant to the current turn, keeping the model more focused and reducing false triggering of distracting slots. As an example, in the MultiWOZ dataset there are 35 slots and 7 domains. Instead of a flat-prediction for all 35 slot statuses simultaneously, by conditioning the slots on domains we only need to consider 5 candidate slots on average. As will be seen in Section~\ref{sec:results}, hierarchical prediction contributes significantly to model quality improvement.

\subsection{Overall procedure}
We summarize the overall model prediction procedure in Algorithm \ref{alg:model}, with the following annotations:

$\mathbf{u}_t$: Utterance at turn $t$.

$\bh^{\{c,n,u\}}$: Vectors of dimension $d$, representations for categorical/non-categorical slots and utterances.

$\bH^{u}$: Matrix of $L_{t}\times d$ where $L_{t}$ is the total number of input tokens up to turn $t$, sequential representations of conversation at time $t$.

$\bH \oplus \bh$: Adding $\bh$ to each row of $\bH$.

$\mathbf{s}^{c,n}$: Name strings for categorical and non-categorical slots.

$\mathcal{C},\mathcal{N}$: Categorical and non-categorical slots sets.

$\mathcal{D}$: Domain set.

$\mathcal{S}_i$: Set of slots belonging to the $i^{th}$ domain.

$\mathbf{q}$: A learnable query vector for domain attention.

$\texttt{Encode}$: Encoding function using pre-trained encoders.

$\texttt{Atten}_{\{d,g,c\}}$: Attention functions for domain, status and categorical slots.

$\texttt{softmax}_{\{d,g, c\}}$: Softmax for domain, status and categorical slots.

$P_{\{s, e\}}$: Position prediction model producing logits at each input token position.

\begin{algorithm}[t] 
\caption{Model prediction procedure}
\begin{algorithmic}[1]
\Procedure{DST}{} 
    \LineComment{Slot encoding}
    \State $\bh^c_i= \texttt{Encode}(\mathbf{s}^c_i), i \in \mathcal{C}$
    \State $\bh^n_i= \texttt{Encode}(\mathbf{s}^n_i), i \in \mathcal{N}$
    
    \LineComment{Context encoding}
    \State $\bh^u_t, \bH^u_t = \texttt{Encode}(\mathbf{u}_1, \ldots, \mathbf{u}_t)$
    
    \LineComment{Domain prediction}
    \State $\ba^d_i = \texttt{Atten}_d(\mathbf{q}, \bH^u_t), i \in \mathcal{D}$
    \State $\mathbf{d}_t = [\texttt{sigmoid}(\ba^d_1), \ldots, \texttt{sigmoid}(\ba^d_{|\mathcal{D}|})]$
    
    \LineComment{Slot status prediction}
    \For{$i \in \mathcal{D}$}
    \If{$\mathbf{d}_t[i] == 1$}
    \For{$s \in \mathcal{S}_i$}
    \State $\ba^g_s = \texttt{Atten}_g(\bh^{\{c, n\}}_s, \bH^u_t)$
    \State $g_s = \texttt{softmax}_g(\bh^{\{c, n\}}_s+\ba^g_s)$
    \EndFor
    \Else
    \State $g_s=\texttt{"inactive"}, s \in \mathcal{S}_i$
    \EndIf
    \EndFor

    \LineComment{Slot value prediction}
    \For{$i \in \mathcal{C}$ and $g_i == \texttt{"active"}$}
    \State $\ba^c_i = \texttt{Atten}_c(\bh^c_i, \bH^u_t)$
	\State $v_i = \texttt{softmax}_c(\bh^u_t + \ba^c_i)$
    \EndFor
    \For{$i \in \mathcal{N}$ and $g_i == \texttt{"active"}$}
    \LineComment{Start/end predictions}
    \State $p^s_i = \argmax\limits_{l=1,\ldots,L_t} P_s(\bH^u_t \oplus \bh^n_i)$
	\State $p^e_i = \argmax\limits_{l=1,\ldots,L_t} P_e(\bH^u_t \oplus \bh^n_i)$
    \EndFor
\EndProcedure
\end{algorithmic}
\label{alg:model}
\end{algorithm}

\section{Training Procedure}

\subsection{Pre-training and fine-tuning}
While pre-training followed by fine-tuning has now almost become a standard training paradigm for NLP tasks, to maximize the efficacy of this procedure it also needs to be properly designed to accommodate the problem structure of the task. Similar to the findings of~\cite{mehri2020dialoglue}, we found two setups that contribute significantly to quality improvement, especially for long conversations as we will show in the experimental section:

\textbf{Domain closeness:} The datasets used for pre-training are preferred to be as close to the dialogue domain as possible. In our work, we pre-train the encoder on a large-scale dataset used to train the Meena open-domain chat bot~\citep{adiwardana2020towards}, whose examples are crawled from social media and internet forums. Although the style of conversations included in the dataset is still very different from task-oriented dialogues, we found it works better than pre-training on the original dataset used by BERT, which is collected from sources of general domains like Wikipedia and Bookcorpus which has little overlap with conversation domains.


\textbf{Continued MLM:} The pre-training procedure endows the model with generic language understanding ability, and following common practice one can directly start fine-tuning the pre-trained model on DST task datasets. A model trained in this way, nevertheless, has missed an opportunity of acquiring task-specific language understanding ability before it learns to make state predictions. Therefore in parallel to tuning for the DST task, we continue the application of the masked language model (MLM) loss as an auxiliary objective during the fine-tuning stage, but on the conversations from both MultiWOZ and ToDBERT~\citep{wu2020tod}. ToDBERT is a collection of public datasets for task-oriented dialogue modeling across multiple domains. ToDBERT matches the MultiWOZ conversation style much better than the Meena dataset used for pre-training, hence it effectively assists the model to gain better understanding in language relevant to the downstream task. The overall size of ToDBERT, however, is tiny compared with the Meena dataset, we therefore only engage this dataset for MLM during fine-tuning. The final loss during the fine-tuning stage is hence the sum of the DST and continued MLM loss.

\subsection{Data Perturbation} \label{sec:perturbation}
To further improve the model's ability of handling long-range conversations, we intentionally perturb the dataset during the fine-tuning stage by extending the length of conversations with randomly inserted utterances. Within each fine-tuning batch, we randomly choose a subset of examples to be perturbed with probability $p$. For each of the chosen examples, we sample $N$ utterances from ToDBERT or MultiWOZ then insert them at random positions of the original conversation as distracting turns. This improves the model's robustness in handling longer conversations by forcing them to extract relevant information from longer, noisier inputs. Alternatively, we may also choose to insert synthetic utterances composed of random words sampled from the vocabulary as perturbation, nevertheless we found this not to be working as effectively as inserting dialogue turns, most likely because random words removes turn fluency, confusing instead of helping the model to improve robustness in dialogue understanding.

\section{Experiments}
\subsection{Setup}
Our experiments are conducted on the MultiWOZ 2.1 dataset. We use the commonly used joint-goal-accuracy (JGA) as metric for DST, defined as the ratio between turns whose all states are correctly predicted and total number of turns. We implemented our approaches using the Lingvo~\cite{shen2019lingvo}\footnote{\href{https://github.com/tensorflow/lingvo}{https://github.com/tensorflow/lingvo}} framework.

\textbf{Pre-training:} For the baseline model, we adopt the standard BERT-Base encoder setup described in~\cite{devlin2018bert, zhang2019find}. The maximum sequence length is set to be 512 tokens for the training examples. For pre-training with the Meena dataset, while we mostly followed the procedure described in~\cite{adiwardana2020towards}, we extended the training sample length to include up to 25 utterances instead of the original 7 utterances. We found using longer examples for pre-training improves quality a little.

\textbf{Fine-tuning:} For fine-tuning with continued MLM as auxiliary loss, we compare two setups: 1) Only apply MLM to MultiWOZ; 2) Apply MLM to both MultiWOZ and ToDBERT datasets. In both setups, the DST model loss is simply added to the MLM loss to form final loss. Note that we excluded MultiWOZ from the ToDBERT dataset during the fine-tuning.

\textbf{Data perturbation:} We compare multiple setups for the perturbation approach introduced in Section~\ref{sec:perturbation}: 1) For source of insertion contents, we consider randomly sampled utterances from ToDBERT, MultiWOZ, or synthetic utterances composed fo random words sampled from the vocabulary; 2) Number of insertion utterances: $N=\{2, 3, 4\}$; 3) Probability of random insertion: $p=\{0.2, 0.4, 0.6\}$; 4) For positions where the sampled utterances are inserted, we consider random boundaries between turns in the original conversation, after only user turns, or agent turns.

\textbf{Baselines:} We compare our results with the following baseline models: 
\newline
\verb|TRADE|: A transferable dialogue state generator that generates dialogue states from utterances using a copy mechanism~\citep{wu2019transferable}.
\newline
\verb|DS-DST|: A dual strategy model that simultaneously handles non-categorical and categorical slots~\citep{zhang2019find}.
\newline
\verb|SST|: A graph attention network that predicts states from utterances and schema graphs containing slot relations in edges~\citep{chen2020schema}.
\newline
\verb|Trippy|: A triple copy strategy where a slot is filled by one of the three copy mechanisms~\citep{heck2020trippy}.
\newline
\verb|SimpleToD|: Training a simple language model that casts DST as a sequence prediction problem~\citep{hosseini2020simple}.
\newline

\subsection{Results}\label{sec:results}
Our main results are presented in Table~\ref{table:results}. The rows of the table correspond to the following setups:
\newline
\verb|Base|: The base model introduced in Section~\ref{sec:base_model}.
\newline
\verb|+hier status|: Enabling hierarchical slot status prediction model in Section~\ref{sec:hier_gate}.
\newline
\verb|+continued MLM|: Apply MLM auxiliary loss on MultiWOZ exampels during fine-tuning.
\newline
\verb|+ToDBERT MLM|: Further add ToDBERT examples for MLM auxiliary loss during fine-tuning.
\newline
\verb|+Meena pre-training|: Instead of pre-training the encoder on the BERT corpus, use the Meena dataset instead. All other setups remain unchanged.
\newline
\verb|+insertion|: Enable data perturbation with insertion introduced in Section~\ref{sec:perturbation}.
\newline\newline
In order to analyze whether these techniques help more on longer or shorter conversations, we in addition report a breakdown of JGA on the shortest and longest 30\% dialogues contained in the MultiWOZ test set, measured by the number of utterances. Among the 7372 turns in the test set, the shortest 30\% consists of dialogues having less or equal to 5 utterances, and the longest 30\% are those having more or equal to 11 utterances.


\begin{table}[t]
\small
\centering
    \centering
    \scalebox{0.95}{
    \begin{tabular}{lccc}
    \hline
    \textbf{Model} & \textbf{All} & \textbf{Short 30\%} & \textbf{Long 30\%}\\
    \hline
    TRADE & 46.00 & - & -\\
    DS-DST  & 52.24 &- &- \\
    SST & 55.23 & - & -\\
    TripPy & 55.29 &- &-\\
    SimpleToD & 56.45 &- &- \\
    \hline
    Base & 48.12 & 70.16 & 20.62\\
    + hier status & 51.81 & 72.74 &  25.33 \\
    + continued MLM & 53.94 & 73.92 & 26.50 \\
    + ToDBERT MLM & 55.65 & 75.42 & 29.50 \\
    + Meena pre-training & 56.64 &75.01 & 30.92 \\
    + insertion & 57.12 & 75.89 & 31.67 \\
    \hline
    \end{tabular}
    }
    \caption{JGA of our methods on the Multiwoz 2.1 benchmark. ``Short 30\%'' and ``long 30\%'' refer to the shortest and longest 30\% of dialogues (measured by the number of turns) in the test set. See text for more explanation.}
    \label{table:results}
\end{table}

\subsubsection{Analysis}\label{sec:analysis}

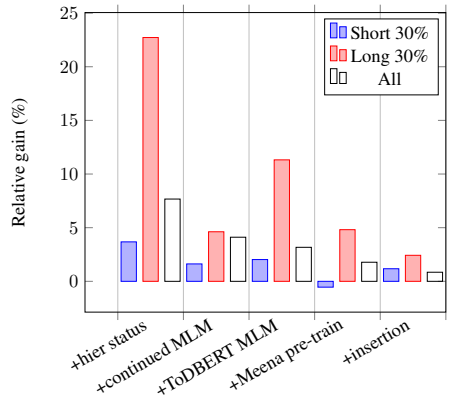
\begin{figure}
\centering
\begin{tikzpicture}[scale=0.70]
\begin{axis}[
	xticklabels={+hier status, +continued MLM, +ToDBERT MLM, +Meena pre-train, +insertion},
	xmax=5,
	xticklabel style={rotate=30,anchor=east},
	ylabel={Relative gain (\%)},
	ybar interval=0.7,
]
\addplot 
	coordinates {(0, 3.677309008) (1, 1.622216112) (2, 2.029220779) (3, -0.5436223813) (4, 1.17317691) (5, 0.)};
\addlegendentry{Short 30\%}
\addplot 
	coordinates {(0, 22.72286822) (1, 4.61902882) (2, 11.32075472) (3, 4.813559322) (4, 2.425614489) (5, 0.)};
\addlegendentry{Long 30\%}		
\addplot[black]
	coordinates {(0, 7.668329177) (1, 4.111175449) (2, 3.170189099) (3, 1.778975741) (4, 0.8474576271) (5, 0.)};
\addlegendentry{All}
\end{axis}

\end{tikzpicture}
\caption{Relative JGA gain on the shortest 30\%, longest 30\% and all conversations.}
\label{fig:rel_gain}
\end{figure}

\textbf{Length vs. improvement:} The main observation from Table~\ref{table:results} is that the overall DST accuracy consistently improves as each of the techniques was introduced, and the gain is more pronounced on longer conversations. To contrast the differences these techniques made on short and long conversations, we plot the relative JGA gain after adopting each technique in Figure~\ref{fig:rel_gain}. It can be seen that each of these techniques, without exception, made more significant improvements on longer conversations than shorter ones, and the biggest relative gain was contributed by enabling hierarchical status prediction.

On the other hand, it can also be seen from Table~\ref{table:results} that while JGA for short conversations can reach more than 75\%, longer ones barely surpassed 31\%. This indicates that further improving long conversations cases are key to the overall DST improvement on benchmarks like MultiWOZ.\\

\noindent\textbf{Accuracy of individual components:} To gain deeper understanding of how individual components of the model, namely categorical slot, non-categorical slot and status predictions, contribute to the overall state tracking behavior, we breakdown the accuracy into individual components for both short and long conversations and report them in Table~\ref{table:breakdown}. From this table we make the following observations:

\begin{enumerate}
    \item Unsurprisingly, accuracy is higher for short conversations than longer ones across the board.
    \item From the simplest base model to the best model at the bottom of table, the accuracy of categorical slots remains almost constant, for both short (98.74\% $\rightarrow$ 98.95\%) and long conversations (98.13\% $\rightarrow$ 98.95\%). The improvement on non-categorical slot accuracy, one the other hand, is much more significant, especially for long conversations (95.9\% $\rightarrow$ 97\% for short, 89.3\% $\rightarrow$ 92.13\% for long). This indicates that categorical slot prediction is more accurate than non-categorical ones, and there is much room for improvement for the latter. It also shows that our proposed techniques mainly improves non-categorical predictions, as conversation becomes longer.
    \item The observation on slot status prediction is similar to that of non-categorical slots. Note however that since the absolute values of status accuracy are very high\footnote{Note that the majority of slots are inactive during a conversation, hence even if a model makes trivial, constant status predictions of ``inactive'', the accuracy can already reach around 90\%.}, further improving status accuracy is very challenging, although the benefit will be huge (as we mentioned in Section~\ref{sec:hier_gate}). For both short and long conversations, the biggest status accuracy improvement is enabled by hierarchical prediction, and the improvement is larger on long conversations.
\end{enumerate}

\begin{table}[t] 
\small
    \centering
    \begin{tabular}{lccc}
    \hline
    \textbf{Model} & \textbf{Cat} & \textbf{Non-cat} & \textbf{Status}\\
    \hline
    Base  & \makecell{98.74 \\ 98.13} & \makecell{95.9 \\ 89.3} & \makecell{98.91 \\ 95.88}\\
    \hline
    + hier gate &  \makecell{98.68 \\ 98.14} & \makecell{95.65 \\ 89.14} & \makecell{99.02 \\ 96.5}\\
    \hline
    + continued MLM & \makecell{98.6 \\ 98.5} & \makecell{95.93 \\ 90} & \makecell{99.06 \\ 96.76}\\
    \hline
    + ToDBERT MLM & \makecell{98.71 \\ 98.45} & \makecell{96.39 \\ 90.73} & \makecell{99.12 \\ 96.9}\\
    \hline
    + Meena pre-training & \makecell{98.9 \\ 98.77} & \makecell{96.47 \\ 91.8} & \makecell{99.12 \\ 96.97} \\
    \hline
    + insertion &\makecell{98.95 \\ 98.95} & \makecell{97 \\ 92.13} & \makecell{99.15 \\ 97.06}\\
    \hline
    \end{tabular}
    \caption{Breakdown of individual model component (categorical and non-categorical slots, status) prediction accuracy. Each row is divided into two sub-rows, corresponding to the shortest and longest 30\% conversations respectively.}
    \label{table:breakdown}
\end{table}

\noindent\textbf{Data perturbation comparison:} We compare the efficacy of different configurations of the perturbation method in Table~\ref{table: perturbation}. Among the variation of perturbation source, probability, number of insertion utterances and positions, the best setup we found is randomly inserting 2 utterances sampled from ToDBERT, with probability 20\%. This is a conservative setup that minimally disturbs the original conversation among our variations, which indicates that while perturbation can be helpful in making models more robust in long conversations, too much perturbation confuses the model and can be counterproductive. We also note that although our end-task is DST for MultiWOZ, sampling utterances from MultiWOZ itself didn't work as well as sampling from ToDBERT. We believe this is because samples from MultiWOZ resemble too much the training example, hence are not as effective in perturbing the model as samples from ToDBERT, a dataset matching the MultiWOZ domain fairly well but not all that similar. \\

\noindent\textbf{Examples of improvement:} To inspect the actual effect our proposed techniques have on the prediction of states, in Appendix~\ref{app:example} we provide several cases for long conversations where errors made by the Meena pre-training model are corrected by the best insertion model.

\begin{table}[t]
\small
    \begin{minipage}{.5\linewidth}
        \centering
        \begin{tabular}{lc}
        \hline
        \textbf{Position} & \textbf{JGA} \\
        \hline
        Random & 57.12\\
        Agent turn only & 56.43 \\
        User turn only & 56.65 \\
        \hline
       \end{tabular}
       \subcaption{Insertion position}
    \end{minipage}
    \begin{minipage}{.5\linewidth}
        \centering
        \begin{tabular}{cc}
        \hline
        \textbf{Probability} & \textbf{JGA} \\
        \hline
        20\% & 57.12 \\
        40\% & 56.73\\
        60\% & 56.66\\
        \hline
       \end{tabular}
       \subcaption{Insertion probability}
    \end{minipage}
    \begin{minipage}{.5\linewidth}
        \centering
        \begin{tabular}{cc}
        \hline
        \textbf{Source} & \textbf{JGA} \\
        \hline
        ToDBERT & 57.12 \\
        Multiwoz & 56.41 \\
        Random & 56.43 \\
        \hline
       \end{tabular}
       \subcaption{Insertion source}       
    \end{minipage}
    \begin{minipage}{.5\linewidth}
    \centering
    \begin{tabular}{cc}
        \hline
        \textbf{Number} & \textbf{JGA} \\
        \hline
        2 & 57.12\\
        3 & 57.02\\
        4 & 56.56\\
        \hline
    \end{tabular}
    \subcaption{Insertion number}
    \end{minipage}
    \caption{Different setups of data perturbation.}
    \label{table: perturbation}
\end{table}

\section{Related Work}
Rapid progress has been made on DST over the past a couple of years. A few key techniques that drove the progress, among others, include state value copying instead of picking from a pre-defined vocabulary~\citep{wu2019transferable}, application of pre-trained BERT encoder~\citep{chao2019bert}, separation between categorical and non-categorical slots~\citep{zhang2019find}, carryover of previously predicted states to avoid repeated prediction via a copy mechanism~\citep{heck2020trippy}, application of large-scale language model for generative stat tracking~\citep{hosseini2020simple}, DST-specific pre-training and fine-tuning procedure~\citep{mehri2020dialoglue}, and various data augmentation approaches~\citep{li2020coco}.

Perhaps mostly related to our work are~\citep{mehri2020dialoglue} and~\citep{li2020coco}, which also adopted domain-specific pre-training and fine-tuning procedures, as well as data augmentation approaches. Both of these works are based on the Trippy model~\citep{heck2020trippy} which introduces recurrence by copying previously predicted values, whereas our models are based on a non-recurrent architecture which repeatedly makes all state predictions. What is more, our analysis mainly focuses on the contrast between the effectiveness of various techniques on short and long conversations, which reveals the importance and larger room for improving longer ones. Our work not only corroborates the existing finding, but also provides additional insights and opportunities for further improvement. 

\section{Conclusion and Future Work}
Improving DST accuracy on longer conversations is a key driving force of the overall quality improvement on benchmarks like MultiWOZ. Our study demonstrated that adopting hierarchical slot status prediction, pre-training on datasets close to the conversation domain, continuation of the masked language model objective during fine-tuning, and data perturbation via utterance insertion all improve DST quality significantly, especially on longer conversations. The findings underscores the importance of proper model design to keep it focused on irrelevant information as a conversation extends, as well as training procedures which enhance the model's ability of understanding longer utterances.

We note however that while we believe our findings are general for DST problems, it is based on a model which repeatedly makes predictions for all states upon receiving every new utterance, which is not a very ideal setup for keeping track of states as the DST task implies a recurrent structure. As next step, we will extend our study to more efficient architectures by introducing such inductive bias for which we believe our findings in the paper will still hold.

\bibliography{iclr2021_conference}
\bibliographystyle{iclr2021_conference}

\appendix\label{app:example}
\section{Appendix}
We show some dialogue examples on which the model before insertion makes the wrong prediction but the model after insertion corrects it. We also provide the slot prediction that the model has corrected. 
\begin{table*}[h]
  \centering
  \begin{tabularx}{\textwidth}{X}
  
    \hline
    User: I am traveling to cambridge and am interested in trying local restaurants and finding a hotel for my stay. \\
    Agent: There are many restaurants. can you please elaborate on what you would like?  \\
    User:  I need to find a hotel with a 3 star rating that includes free wifi. \\
    Agent: I have 5 options for you, located all over town. do you have a certain area or price range in mind?\\
    User: I want Chinese food in cheap price range in \textbf{west} side of town. a 3 star hotel that is expensive and includes wifi. Also the hotel address, area, and postcode please.\\
    \textbf{Before insertion}: restaurant-area = `'\\
    \textbf{After insertion}:  restaurant-area = `west'\\
    
    \hline
    
    User: I would a train. I'm going from birmingham new street and it needs to arrive by 11:00.\\
    Agent: What day would you like to travel? \\
    User: I am leaving on Saturday, and the train should go to Cambridge.  \\
    Agent: tr8259 will arrive in cambridge at 10:23. Would you like me to book a ticket for you on that train? \\
    User: Not yet. First i would like to get price and departure time. \\
    Agent: It will depart at 07:40 and it will cost 60:08 pounds. do you need help with anything else? \\
    User: And I need a place to stay. \\ 
    Agent: Okay there are many options. Do you have a price preference or area? \\
    User: I would like something in the east. I also prefer a guest house. \\ Agent: There is \textbf{a and b guest }, it is in the east part of town
    User: That sounds good, I have 7 people in total. \\
    \textbf{Before insertion}: hotel-name = `'\\
    \textbf{After insertion}: hotel-name = `a and b guest'
    \\
    \hline
    
    User: i need a train to cambridge, departing from the peterborough station.
   \\
   Agent: There are a total of 266 trains traveling from peterborough to cambridge. what day would you like to leave?
  \\
  User:I would like to travel on saturday and arrive by 15:15. 
  \\
  Agent: i have a train scheduled to depart peterborough at \textbf{14:19} and will arrive in cambridge by 14:38. would you me like to book this for you? \\
  
  User: Yes, for 2 people, please.\\
  \textbf{Before Insertion}: train leave at = `15:15' \\
    \textbf{After Insertion}: train leave at = `14:19'\\
  
  \hline

  \end{tabularx}
  \caption{Dialogues where the model with insertion has corrected the wrong prediction.}
  \label{tab:dialog_one}
\end{table*}

\end{document}